\documentclass[10pt,twocolumn,letterpaper]{article}

\usepackage[pagenumbers]{cvpr} 
\usepackage{times}
\usepackage{epsfig}

\usepackage{arydshln}
\usepackage{amssymb}
\usepackage{multirow}
\usepackage{float} 
\mathchardef\mhyphen="2D
\usepackage[font=small,skip=1pt]{caption}
\usepackage{graphicx}
\usepackage{amsmath}
\usepackage{amssymb}
\usepackage{booktabs}

\usepackage[pagebackref,breaklinks,colorlinks]{hyperref}

\usepackage[capitalize]{cleveref}
\crefname{section}{Sec.}{Secs.}
\Crefname{section}{Section}{Sections}
\Crefname{table}{Table}{Tables}
\crefname{table}{Tab.}{Tabs.}

\def\cvprPaperID{11215} 
\def\confName{CVPR}
\def\confYear{2023}

\begin{document}

\title{Enhancing Self-Training Methods}

\author{Aswathnarayan Radhakrishnan$^1$ \hspace{0.5cm} Jim Davis$^1$ \hspace{0.5cm} Zachary Rabin$^1$ \\ Benjamin Lewis$^2$ \hspace{0.5cm} Matthew Scherreik$^2$ \hspace{0.5cm} Roman Ilin$^2$ \\ \\
$^1$\ Department of Computer Science and Engineering \\ Ohio State University \\
{\tt\small \{radhakrishnan.39,\ davis.1719,\ rabin.30\}@osu.edu}\\\\
$^2$\ AFRL/RYAP, Wright-Patterson AFB \\
{\tt\small \{benjamin.lewis.13,\ matthew.scherreik.1,\ roman.ilin.1\}@us.af.mil}
}
\maketitle

\begin{abstract}
Semi-supervised learning approaches train on small sets of labeled data along with large sets of unlabeled data. Self-training is a semi-supervised teacher-student approach that often suffers from the problem of ``confirmation bias" that occurs when the student model repeatedly overfits to incorrect pseudo-labels given by the teacher model for the unlabeled data. This bias impedes improvements in pseudo-label accuracy across self-training iterations, leading to unwanted saturation in model performance after just a few iterations. In this work, we describe multiple enhancements to improve the self-training pipeline to mitigate the effect of confirmation bias. We evaluate our enhancements over multiple datasets showing performance gains over existing self-training design choices. Finally, we also study the extendability of our enhanced approach to Open Set unlabeled data (containing classes not seen in labeled data). 
\end{abstract}

\section{Introduction}

In today's data-driven world, deep learning techniques have become the predominant approach for computer vision tasks (such as image classification and object detection). Most state-of-the-art (SOTA) deep learning models use large-scale labeled datasets (e.g., ImageNet \cite{Deng2009a}, JFT-3B \cite{Zhai2022a}, Instagram-3.5B \cite{Mahajan2018a}), a few of which are proprietary and cannot be leveraged by the public. It is challenging in practice to curate and annotate large labeled real-world datasets across different data domains and learning tasks. However, it is much easier to collect large quantities of \emph{unlabeled data} in real-world domains (e.g., remote sensing imagery \cite{Radhakrishnan2019a, Motlagh2021a}, medical imagery \cite{Haque2021a, Tajbakhsh20201a}). Semi-supervised learning (SSL) techniques are designed to jointly leverage small \emph{labeled} datasets along with large \emph{unlabeled} datasets to improve model performance.

\begin{figure}
\begin{center}
\includegraphics[width=8cm]{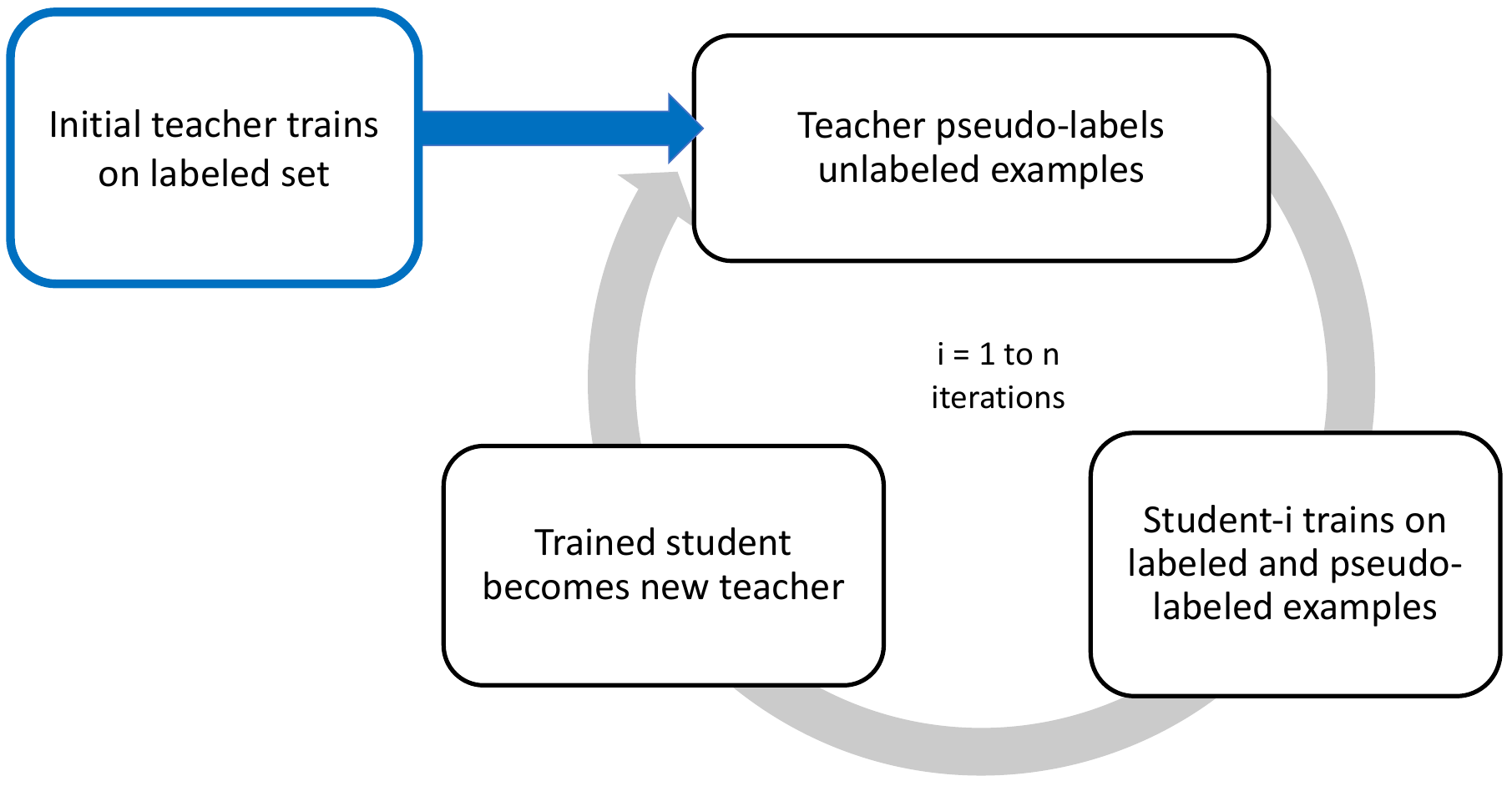}
\end{center}
   \caption{Basic iterative self-training pipeline.}
\label{fig:basic}
\end{figure}

Self-training (ST) \cite{Scudder1965a, Yarowsky1995a, Riloff1996a, Riloff2003a} is an iterative SSL method where a ``teacher" model trained on the labeled data annotates the unlabeled data with pseudo-labels. The subsequent learning of the ``student" model uses both the labeled and pseudo-labeled data. This process is iterated, as shown in Fig.~\ref{fig:basic}. The major caveat of pseudo-labeling is the introduction of noisy pseudo-labels from incorrect predictions by the teacher. These noisy pseudo-labels accumulate over time resulting in the model developing a bias toward incorrectly predicted pseudo-labels. This issue is known as the ``confirmation bias" problem \cite{Arazo2020a}. 

SSL techniques that learn from limited labeled data employ consistency regularization techniques \cite{Laine2017a, Li2020a, Tarvainen2017a} to reduce confirmation bias. Another popular method for reducing confirmation bias when enough labeled data is available is the NoisyStudent (NS) \cite{Xie2020a} pseudo-labeling approach that uses softmax confidence thresholding to filter out under-confident pseudo-label predictions. This approach also found that training a student model larger than the initial teacher made the student more robust to handle noisy pseudo-labels. To reduce confirmation bias, we explore multiple design choices and variations to the NS iterative learning pipeline. 

In this paper, we analyze existing methods and propose additional novel modifications that include using calibrated teacher models, entropy-based pseudo-label thresholding, and custom SplitBatch sampling. The proposed enhancements are modular and can be adapted to work with multiple existing ST methods. We demonstrate the use of the modifications to enhance ST across multiple benchmark datasets.  Lastly, we present a practical scenario using real-world Open Set unlabeled data that contains both data belonging to the target training classes and data from additional/unwanted classes. We demonstrate our enhanced ST technique using an Open Set recognition approach integrated with our ST pipeline to improve performance even when trained with challenging Open Set data.


\section{Related Work}
SSL is an active field of research in deep learning \cite{Chapelle2009a, Zhu2005a, Ouali2020a}. Consistency regularization and pseudo-labeling are some of the most commonly used SSL methods to learn from large sets of unlabeled data over recent years.

Consistency regularization approaches \cite{Sajjadi2016a, Laine2017a, Tarvainen2017a, Athiwaratkun2019a} follow the data manifold assumption that perturbations applied to the training data, such as data augmentation, should not increase the likelihood of the predicted target labels switching classes. These methods minimize the difference in predictions between an unlabeled training sample and its perturbed counterpart. In \cite{Xie2020a}, they discussed that consistency regularization methods work better in scarcely labeled data scenarios as they simultaneously learn to generate target predictions while maintaining the consistency requirements described above. However, pseudo-labeling methods are preferred when the given labeled data is sufficient to train a fully-supervised model for generating quality pseudo-labels (unlike with consistency regularization methods). In our work, we focus on such pseudo-labeling SSL methods.

The ST pseudo-labeling approach is one of the oldest and most widely used SSL approaches. In the NS \cite{Xie2020a} approach, the initial teacher model is trained on labeled data and then used to generate pseudo-labels for the unlabeled data. The student model is trained on both datasets (labeled + pseudo-labeled). They also found that injecting model noise (Dropout \cite{Srivastava2014a}) and data noise (RandAugment \cite{Cubuk2020a}) into the student training made it more robust to noisy pseudo-labels. NS also employed a student model larger than the original teacher to improve generalization in their experiments. They iterated the above steps using the student model from the previous iteration as the new teacher model, as shown in Fig.~\ref{fig:basic}. We adapt this popular iterative teacher-student ST pseudo-labeling pipeline in our work.


There have been various approaches introduced to help reduce confirmation bias \cite{Arazo2020a, Su2021a} in SSL methods. In \cite{Grandvalet2004a}, a regularization term was added to the loss function to encourage the model to make confident low-entropy pseudo-label predictions. In \cite{Xie2020a}, they used handpicked softmax thresholds on pseudo-labels to filter out noisy low-confidence predictions that can increase confirmation bias. In \cite{Li2020a}, a Gaussian mixture model was applied to divide the training data dynamically into clean and noisy sets using a per-sample loss distribution. They trained two networks where each network used the other network's divided set to reduce confirmation bias. In \cite{Yang2022a}, handpicked confidence thresholding was used to separate the unlabeled data into clean in-distribution and noisy out-of-distribution data. They applied a class-aware clustering module for the in-distribution pseudo-labeled data along with a contrastive learning module to mitigate the noise in the out-of-distribution pseudo-labeled data. Our work proposes dataset-adaptive thresholding methods to filter out noisy pseudo-labels instead of manually fine-tuning handcrafted techniques for each dataset.

In \cite{Arazo2020a}, substantial data augmentation and regularization policies such as RandAugment \cite{Cubuk2020a},  Mixup \cite{Zhang2018a}, and Dropout \cite{Srivastava2014a} were shown to minimize the effect of confirmation bias. Following \cite{Arazo2020a}, we apply an improved dynamic version of Mixup called SAMix \cite{Li2021a} in this work to help minimize confirmation bias. SAMix learns to saliently mix two images from the dataset within the training process, adapting the mixing policy without the need for tuning the Mixup hyperparameters. Most of the above methods require complex modifications to the training architecture and optimization strategies to mitigate the problem of confirmation bias. Overall, we propose modular enhancements to fundamental training components that can adapt to existing ST pipelines to improve SSL performance.

\section{Design Choices within Self-Training}
We aim to enhance the baseline NS approach by modifying different stages of the pipeline to generate and select better pseudo-labels (having higher pseudo-label accuracy) to help mitigate confirmation bias. We first introduce the notation used in our pipeline. Let the training data $\mathit{D}$ be composed of the labeled subset with pairs $\mathit{D_{l}} = \left \{ \left ( x_{i}, y_{i} \right ) \right \}_{i=1}^{N_{l}}$ where $x_{i}$ denotes a labeled sample (e.g., image) and $y_{i}$ denotes its corresponding ground truth label. The unlabeled subset contains data $\mathit{D_{u}} = \left \{ \left ( \tilde{x}_{i} \right ) \right \}_{i=1}^{N_{u}}$ with no labels. A pseudo-label predicted for an unlabeled sample $\tilde{x}_{i}$ will be denoted as $\tilde{y}_{i}$. Let $\mathit{f_{T}}$ and $\mathit{f_{S}}$ denote the teacher and student models, respectively. We now propose the following comparisons of possible design choices to the fundamental components in the ST pipeline.

\subsection{Hard vs. Soft Loss}
Supervised deep learning models trained with clean one-hot ground truth labels generally employ a \emph{categorical} cross-entropy loss known as a \textbf{hard loss} (using a known, single ground truth label for each example and associated softmax prediction). ST techniques must handle both clean ground truth labels and noisy pseudo-labels generated from the softmax prediction vectors of the teacher model on unlabeled data. Previous work \cite{Xie2020a, Arazo2020a} has shown that using a \textbf{soft loss} (used in NS) with the \emph{entire} softmax vector of pseudo-label predictions as targets in a full cross-entropy loss works better than a hard loss, with a softmax distribution over all training classes, helping to reduce overfitting to noisy targets. 

\subsection{Student Initialization}
ST is an iterative technique that trains a new student model for every iteration. Two main methods exist for initializing the student model for each ST iteration. \textbf{Fresh-training} (used in NS) initiates every student model from scratch (i.e., newly initialized model weights). Conversely, \textbf{fine-tuning} uses the weights of the model from \emph{any} previous iterations with the best accuracy (on validation data) to initialize the student model and then fine-tune the weights during the current iteration. 

\subsection{Labeled/Pseudo-labeled Mini-Batch}
Student models in ST learn from the labeled ground truth subset and the unlabeled subset with noisy pseudo-labels. ST commonly uses a \textbf{randomly collected (uniform) mini-batch} (used in NS) that tends to overfit due to their bias towards selecting a larger number of noisy pseudo-labeled than labeled data during training, as the unlabeled subset sizes are usually much larger than the labeled subset. We propose a \textbf{custom SplitBatch} approach that collects a user-specified split of labeled and pseudo-labeled examples for every mini-batch. Our approach uses bootstrapping to select examples from the limited labeled subset containing only clean labels, providing an additional regularization effect to counter the noise from the pseudo-labels. The user controls the hyperparameter that sets the ratio of labeled to pseudo-labeled examples in a mini-batch, making the approach adaptive to differently sized labeled/pseudo-labeled subsets. For example, a larger ratio of labeled to pseudo-labeled examples can be used for datasets having a large number of labeled samples. 

With a split batch of labeled and pseudo-label examples, the loss function should similarly engage a split loss. In this work, we combine a labeled loss $L_{lab}$ and a pseudo-labeled loss $L_{pslab}$ with \emph{equal} contributions into a custom $\textrm{MixedLoss}$ function 
\begin{equation}
    L_{mix} = \lambda _{b}L_{lab}  + (1-\lambda _{b})L_{pslab}
\label{eqn2}
\end{equation}
where $\lambda _{b}$ is a hyperparameter set to 0.5 to balance the loss between the labeled and pseudo-labeled examples in all of our experiments.

\subsection{Sampling Techniques}
ST methods can easily generate class-unbalanced and low-confidence pseudo-labeled subsets that can also increase confirmation bias. NS uses a \textbf{na\"ive softmax-thresholded class balancing} technique that first uses the uncalibrated softmax scores of pseudo-label predictions to threshold high-confidence predictions (softmax scores for the argmax class \textgreater~0.3). NS then samples a user-specified number of thresholded pseudo-labeled examples that have the highest softmax confidence across every class, oversampling examples from classes not having enough pseudo-labeled examples (less than the user-specified count per-class). This method requires a handcrafted softmax threshold and per-class sampling count for every dataset. 

Our previously defined \textbf{SplitBatch sampler} can be adapted to also dynamically re-weight and sample pseudo-labeled examples using two different sample weightings. The first weighting uses inverted per-class lengths ($\frac{1}{N_{c}}$ where $N_{c}$ is the number of pseudo-labeled examples belonging to class $c$). This method assigns larger weights to classes with a lower number of pseudo-labels which thus will be oversampled during training. The second set of sample weightings uses the per-class normalized softmax confidence scores
\begin{equation}
    normalizedSoftmax = \frac{max (\tilde{y})}{max(S_{c})}
\label{eqn0.5}
\end{equation}
where $\tilde{y}$ is the complete pseudo-label softmax vector prediction by the teacher model having argmax predicted class $c$ for a given unlabeled data sample $\tilde{x}$, and $S_{c} = \{max(\tilde{y_{1}}), ..., max(\tilde{y}_{N_{c}})\}$ is the set of max softmax scores for all pseudo-label predictions belonging to class $c$.
These normalized softmax scores scale the weights per-class to avoid oversampling from classes with higher softmax confidence. We average the two sampling weights  (class-length and the normalized softmax-confidence-based weights) and distribute them across the unlabeled examples to dynamically obtain the final sampling weights for every dataset without the need for handcrafted thresholding and class-balancing.


\subsection{Pseudo-Label Selection}
NS uses the \textbf{na\"ive softmax thresholding} approach (described above), employing softmax scores as a metric to determine pseudo-label confidence. However, modern deep neural networks are known to be poorly calibrated \cite{Guo2017a}, implying that the softmax prediction probabilities do not accurately represent the true likelihood of the predictions. Hence, the uncalibrated softmax score is a poor confidence metric for rejecting noisy samples and thus can increase confirmation bias. 

Alternatively, we propose adding a temperature-scaling calibration \cite{Guo2017a} step in the ST pipeline to the current teacher model for generating calibrated pseudo-label softmax predictions. We use a grid search over 400 linearly spaced temperature values between 0.05 and 20 and choose the optimal value, denoted by $\tau $, with the lowest Expected Calibration Error \cite{Naeini2015a} on the validation data. We then apply $\tau $ to soften/sharpen the softmax pseudo-label predictions of the teacher model to get a full softmax vector of pseudo-labels 
\begin{equation}
    \tilde{y} = softmax\left ( \frac{\mathit{f_{T}}\left ( \tilde{x} \right )}{\tau} \right )
\label{eqn1}
\end{equation}
where $\mathit{f_{T}}\left ( \tilde{x} \right )$ denotes the output logits of the teacher model for a given unlabeled sample $\tilde{x}$ and $\tilde{y}$ is the calibrated pseudo-label softmax vector. 

We next propose using \textbf{entropy thresholding} of the calibrated softmax pseudo-label vector rather than simply thresholding the softmax score for the argmax class to determine if the pseudo-label is acceptable. We calculate the \textbf{normalized entropy} (dividing by $log \left ( N_{c} \right )$ for $N_{c}$ classes) of the calibrated pseudo-labels of validation data and then run a grid search over 500 threshold values between 0 and 1. We then calculate the true-positive rate (TPR) and false-positive rate (FPR) for each of the possible entropy thresholds on the validation data. We perform ROC analysis \cite{Fawcett2006a} by plotting the TPR against the FPR at the various thresholds and selecting the optimal threshold with the lowest Euclidean distance to the top left corner (optimal/perfect classification) of the ROC curve. This method for pseudo-label selection can dynamically adapt to different datasets, unlike the na\"ive approach of using handpicked softmax thresholds for each dataset.

\subsection{Teacher Size}
Lastly, the NS approach uses a \textbf{smaller-sized initial teacher model} trained on the clean labels and a larger student model (and thus a larger subsequent teacher) trained jointly on the labeled and pseudo-labeled examples. As previously mentioned, the NS student model incorporates model noise (Dropout) and data noise (strong data augmentation techniques). A natural alternative to their approach is a \textbf{same-sized teacher-student model}, where the teacher and student have the same model size but use \emph{stronger} data augmentation techniques (SAMix+RandAugment). We compare both small and large same-sized teacher-student models. 

Given the above-listed design choice alternatives, we now compare them in an experimental setting to enhance the basic ST pipeline. 

\begin{table*}[t]
\centering
\begin{tabular}{|l|l|}
\hline
Experiment                                    & Description                                                            \\ \hline\hline
Exp 1. Hard vs. Soft Loss                          & One-Hot Categorical Cross-Entropy Loss vs. Soft Cross-Entropy Loss     \\ 
Exp 2. Student Initialization          & Training from Scratch vs. Fine-tuning Student Iterations               \\ 
Exp 3. Labeled/Pseudo-labeled Mini-Batch                    & Random Mini-batch (Mixed) vs. SplitBatch (Labeled + Pseudo-labeled) \\ 
Exp 4. Sampling Techniques & Na\"ive Softmax-Thresholded Class Balancing vs. Weighted SplitBatch Sampling                    \\ 
Exp 5. Pseudo-Label Selection                       & Na\"ive Softmax Thresholding vs. Calibrated Entropy Thresholding                                         \\ 
Exp 6. Teacher Size                            & Smaller vs. SameSized Teacher                                          \\ \hline
\end{tabular}
\caption{Experimental comparison roadmap.}
\label{Table0}
\end{table*}

\section{Experiments}
We aim to create an improved ST model by exploring the previously described design choices using the sequential strategy used by \cite{Liu2022a}, where a linear series of experiments are employed to modernize a baseline model by augmenting the model with the best component obtained after each design choice comparison. Similarly, we start from the basic ST iterative learning pipeline, follow the roadmap described in Table \ref{Table0}, and choose the best design choices sequentially using a majority voting selection across multiple benchmark datasets to create an enhanced ST approach (rather than evaluating all possible combinations of design choices).  We designed the order of experiments in the roadmap, starting from fundamental components (such as the loss function) and moving toward finer settings (such as sampling techniques and model sizes). 


\subsection{Datasets}

We created custom labeled/unlabeled subsets from various benchmark datasets (SVHN \cite{Netzer2011}, CIFAR-10 \cite{Krizhevsky2009a}, and CIFAR-100 \cite{Krizhevsky2009a}) following the standard subset sizes from previous SSL work \cite{Arazo2020a}, as shown at the top of Table \ref{Tabledata}. For each dataset, we also created a validation subset with ground-truth labels for hyperparameter tuning and evaluating model performance during training, and a corresponding test subset for evaluating model inference. We first evaluated the experiments described in Table \ref{Table0} on the three datasets and constructed the enhanced approach using the best component choices. We further evaluated the generalization performance of the resulting enhanced approach with different labeled/unlabeled dataset splits and model sizes on additional datasets (CINIC-10 \cite{Darlow2018a}, TinyImageNet \cite{letiny}) as shown at the bottom of Table \ref{Tabledata}. We also compared the performance of our resulting enhanced approach with the original NS method on all the datasets. Note that every dataset except SVHN is class-balanced. Finally, we extended the enhanced approach with a basic Open Set detection technique to help filter out (suppress) additional/unwanted classes in a custom-built Open Set version of CIFAR-10/CIFAR-100 with 110 separate classes.

\subsection{Comparison Roadmap}
We trained a supervised baseline model for each dataset on only the labeled data subset for three runs initialized with different random seeds for each experiment in the roadmap (Exp.~1 to 6). We reported the best mean test set score obtained across three student iterations (unless otherwise mentioned for experiments below). We used a ResNet(R)-18 \cite{He2016a} for SVHN and a WideResNet(WRN) 28-2 \cite{Zagoruyko2016WideRN} + SAMix for the CIFAR datasets, unless otherwise mentioned. SAMix was not used on SVHN as certain mixing data augmentation policies were expected not to be appropriate for digit classification datasets (e.g., crops, flips). We also applied RandAugment to each model/dataset with the hyperparameter settings for each dataset given in the original work \cite{Cubuk2020a}. RandomCrop and RandomHorizontalFlip  were included in the data augmentation policy for the CIFAR datasets only. We trained the initial teacher model for 400 epochs on the labeled subsets of all datasets following the suggested training and hyperparameter settings for SAMix \cite{Li2020a}. Each student iteration was trained for 100 epochs (which actually includes more mini-batches per-epoch than the teacher). We used a batch size of 100 for all experiments. Table \ref{Table1} shows the mean initial teacher accuracy trained only using the labeled subset (note that these scores are expected to be lower than fully-supervised SOTA benchmarks that use the complete datasets). \\

\begin{table}[t]
\centering
\begin{tabular}{|c|c|c||c|c|c|}
\hline 
Dataset (NumClasses) & Lab & ULab & Val & Test \\ \hline \hline
SVHN (10)             & 1K          & 70K   & 1K  & 26K  \\
CIFAR-10 (10)      & 4K           & 42K   & 4K  & 10K  \\
CIFAR-100 (100)     & 10K     & 30K   & 10K & 10K  \\
\hdashline
CINIC-10 (10)      & 1K or 20K  & 150K  & 10K & 90K  \\
TinyImageNet (200)     & 20K      & 60K   & 20K & 10K  \\ \hline
\end{tabular}
\caption{Dataset sizes. (Lab: Labeled, ULab: Unlabeled, Val: Validation, Test: Test set sizes)}
\label{Tabledata}
\end{table}

\begin{table}[t]
\centering
\begin{tabular}{|c|c|c|}
\hline
Datasets  & Models  & Mean Teacher Top-1 Acc \\
\hline\hline
SVHN 1K      & R18     & 75.69           \\
CIFAR-10 4K  & WRN28-2 & 83.66           \\
CIFAR-100 10K & WRN28-2 & 63.9    \\  
\hline
\end{tabular}
\caption{Labeled subset supervised baseline results.}
\label{Table1}
\end{table}

\noindent \textbf{Exp 1. Hard vs. Soft Loss} \\
Table \ref{Table2} shows the results of the comparison between ST models using \textbf{soft} loss vs. \textbf{hard} loss. We can see that soft loss employed by the NS approach performs better on SVHN and CIFAR-10. Both losses degrade the performance on CIFAR-100 from the supervised baseline because the basic ST pipeline can easily overfit to the noisier CIFAR-100 pseudo-labels (CIFAR-100 has the worst initial teacher in Table \ref{Table1}, which would generate the noisiest pseudo-labels). \textit{By 2-1 majority vote, we apply the \textbf{soft loss} (cross-entropy loss with soft targets) henceforth in our experiments.} \\

\begin{table}[t]
\centering
\begin{tabular}{|c|c|cc|}
\hline
\multirow{2}{*}{Datasets} & \multirow{2}{*}{Models} & \multicolumn{2}{c|}{Mean Student Top-1 Acc}     \\  
                          &                         & \multicolumn{1}{c}{Hard Loss} & Soft Loss      \\ \hline\hline
SVHN                      & R18                     & \multicolumn{1}{c|}{80.96}     & \textbf{81.22*} \\
CIFAR-10                  & WRN28-2                 & \multicolumn{1}{c|}{85.15}     & \textbf{86.85*} \\
CIFAR-100                 & WRN28-2                 & \multicolumn{1}{c|}{59.24}     & \textbf{62.24}    \\
\hline
\end{tabular}
\caption{Hard vs. Soft loss results. (\textbf{Bold}: Best result in table,\\ \textbf{*}: Current best result for each dataset)}
\label{Table2}
\end{table}

\noindent \textbf{Exp 2. Student Initialization} \\
We next evaluated \textbf{fresh-training} vs. \textbf{fine-tuning} of ST models across training iterations. Table \ref{Table3} shows that fine-tuning improved ST performance across all datasets compared to the fresh-training approach used in NS. The performance gains provided by fine-tuning show that carrying over the learned weights of the teacher model during ST is beneficial. \textit{Hence we use \textbf{fine-tuning} for the remaining set of experiments.}  \\
\begin{table}[t]
\centering
\begin{tabular}{|c|c|cc|}
\hline
\multirow{2}{*}{Datasets} & \multirow{2}{*}{Models} & \multicolumn{2}{c|}{Mean Student Top-1 Acc}     \\  
                          &                         & \multicolumn{1}{c}{Fresh-Train} & Fine-Tune     \\ \hline\hline
SVHN                      & R18                     & \multicolumn{1}{c|}{81.22}     & \textbf{81.55*} \\
CIFAR-10                  & WRN28-2                 & \multicolumn{1}{c|}{86.85}     & \textbf{87.45*} \\
CIFAR-100                 & WRN28-2                 & \multicolumn{1}{c|}{62.24}     & \textbf{65.86*}        \\
\hline
\end{tabular}
\caption{Student initialization comparison results. (\textbf{Bold}: Best result in table, \textbf{*}: Current best result for each dataset)}
\label{Table3}
\end{table}

\noindent \textbf{Exp 3. Labeled/Pseudo-labeled Mini-Batch} \\
Table \ref{Tabel4} shows the results of using the default \textbf{random mini-batch} approach used in NS against our proposed \textbf{SplitBatch} approach and the associated MixedLoss function $L_{mix}$. For SVHN and CIFAR-10, which have small amounts of labeled examples, we used a 20/80\% labeled/pseudo-labeled batch split, whereas, for CIFAR-100, we used a 40/60\% split as it has a larger number of labeled examples. Our proposed custom SplitBatch approach performed slightly better by oversampling (sampling with replacement) labeled examples in every mini-batch to prevent overfitting to noisy pseudo-labels. \textit{We use the better-performing custom \textbf{SplitBatch approach} along with the same split percentages used for each dataset going forward.}\\
\begin{table}[t]
\begin{tabular}{|c|c|cc|}
\hline
\multirow{2}{*}{Datasets} & \multirow{2}{*}{Models} & \multicolumn{2}{c|}{Mean Student Top-1 Acc}     \\  
                          &                         & \multicolumn{1}{c}{Random} & SplitBatch (L/PS\%)     \\ \hline\hline
SVHN                     & R18                     & \multicolumn{1}{c|}{81.55}     & \textbf{81.68*} (20/80\%)\\
CIFAR-10               & WRN28-2                 & \multicolumn{1}{c|}{87.45}     & \textbf{87.61*} (20/80\%) \\
CIFAR-100                 & WRN28-2                 & \multicolumn{1}{c|}{65.86}     & \textbf{65.90*} (40/60\%)        \\
\hline
\end{tabular}
\caption{Labeled/Pseudo-labeled mini-batch collection comparison results. (\textbf{Bold}: Best result in table, \textbf{*}: Current best result for each dataset, L/PS\%: Labeled/Pseudo-labeled split percentage)}
\label{Tabel4}
\end{table}

\noindent \textbf{Exp 4. Sampling Techniques} \\
We compared the \textbf{na\"ive softmax-thresholded class balancing} approach employed by NS with our proposed \textbf{\underline{weighted} SplitBatch sampler} (employs class-length balancing and confidence weighting using the same splits from the previous experiment). The student iterations henceforth are trained for 150 epochs (instead of 100) as these methods need more epochs to converge as they work on thresholded/oversampled pseudo-labeled subsets (previous experiments used the complete set of pseudo-labeled data). Unlike the original NS approach, which used a softmax threshold of 0.3 on 1000-class ImageNet, we used a larger threshold of 0.5 in this experiment to threshold noisy pseudo-labeled data as the maximum number of classes is only 100 in our datasets, resulting in higher softmax values for the argmax classes (0.5 is also a natural decision boundary between low and high confidence). In this experiment, we employed our weighted SplitBatch sampling without any thresholding of pseudo-labeled data. Table \ref{Table5} shows that our weighted SplitBatch sampler performed better on the CIFAR datasets. In contrast, the na\"ive method, which uses confidence-sorted sampling, is better on SVHN that had more highly confident pseudo-labels. However, the na\"ive method samples more incorrect high-confidence pseudo-labels on CIFAR datasets than SVHN, leading to performance degradation on CIFAR data. \textit{Henceforth, we employ our \textbf{weighted SplitBatch sampler} in the ST pipeline.}  \\
\begin{table}[t]
\centering
\begin{tabular}{|c|c|cc|}
\hline
\multirow{2}{*}{Datasets} & \multirow{2}{*}{Models} & \multicolumn{2}{c|}{Mean Student Top-1 Acc}     \\  
                          &                         & \multicolumn{1}{c}{Na\"ive} & W. SplitBatch     \\ \hline\hline
SVHN                      & R18                     & \multicolumn{1}{c|}{\textbf{83.30*}}     & 81.07 \\
CIFAR-10                  & WRN28-2                 & \multicolumn{1}{c|}{86.56}     & \textbf{87.95*} \\
CIFAR-100                 & WRN28-2                 & \multicolumn{1}{c|}{68.47}     & \textbf{69.53*}        \\
\hline
\end{tabular}
\caption{Sampling techniques comparison results. (Na\"ive: na\"ive softmax-thresholded class balancing, W. SplitBatch: Weighted SplitBatch Sampler, \textbf{Bold}: Best result in table, \textbf{*}: Current best result for each dataset)}
\label{Table5}
\end{table}


\noindent \textbf{Exp 5. Pseudo-Label Selection} \\
We compared the same \textbf{na\"ive softmax-thresholding} approach used by NS from the previous experiment with our enhanced approach employing \textbf{calibrated entropy thresholding} for pseudo-label selection along with our weighted SplitBatch sampling. Table \ref{Table7} shows that our enhanced approach performed better than na\"ive softmax-thresholding on all datasets. We also improved upon the SVHN na\"ive sampling scores from the previous experiment, demonstrating the efficacy of using our weighted SplitBatch sampler and calibrated entropy thresholding in tandem. \textit{Hereafter, we apply our \textbf{calibrated entropy thresholding} method.}\\

\begin{table}[t]
\centering
\begin{tabular}{|c|c|cc|}
\hline
\multirow{2}{*}{Datasets} & \multirow{2}{*}{Models} & \multicolumn{2}{c|}{Mean Student Top-1 Acc}     \\  
                          &                         & \multicolumn{1}{c}{Softmax} & Entropy     \\ \hline\hline
SVHN                      & R18                     & \multicolumn{1}{c|}{83.30}     & \textbf{84.28*} \\
CIFAR-10                  & WRN28-2                 & \multicolumn{1}{c|}{86.56}     & \textbf{88.52*} \\
CIFAR-100                 & WRN28-2                 & \multicolumn{1}{c|}{68.47}     & \textbf{69.84*}       \\
\hline
\end{tabular}
\caption{Pseudo-label selection comparison results. (\textbf{Bold}: Best result in table, \textbf{*}: Current best result for each dataset)}
\label{Table7}
\end{table}

\noindent \textbf{Exp 6. Teacher Size} \\
Table \ref{Table8a} shows the results of employing \textbf{differently sized teacher models} in the ST pipeline. We found that the NS approach of using a smaller teacher (R18 for SVHN and WRN28-2 for CIFAR-10 and CIFAR-100) and a larger student (R34 for SVHN and WRN40-2 for CIFAR-10 and CIFAR-100) with model noise (Dropout) is unnecessary once we add our selected design choices from previous experiments to reduce confirmation bias. The results in Table \ref{Table8a} compare the NS approach with the SmallerSameSized (SSS: using the \emph{smaller} teacher model size as the student model size) and LargerSameSized (LSS: using the \emph{larger} teacher model size as the student model size) approaches. We can see that SSS models performed on par on SVHN and slightly better on CIFAR than the NS approach, whereas the LSS approach improved accuracy across all datasets. \textit{Hence we use \textbf{larger same-sized teacher-student models} in our ST pipeline.} \\

\begin{table*}[ht!]
\centering
\begin{tabular}{|c|c|ccc|}
\hline
\multirow{2}{*}{Datasets} & \multirow{2}{*}{Models}                         & \multicolumn{3}{c|}{Mean Student Top-1 Acc}                                                                    \\ 
                          &                                                 & \multicolumn{1}{c}{NS} & \multicolumn{1}{c}{SSS} & \multicolumn{1}{l|}{LSS} \\ \hline\hline
SVHN                      & NS: R18+R34, SSS: R18, LSS: R34                 & \multicolumn{1}{c|}{84.28}         & \multicolumn{1}{c|}{84.28} & \textbf{86.71*}                      \\ 
CIFAR-10                  & NS: WRN28-2+WRN40-2, SSS: WRN28-2, LSS: WRN40-2 & \multicolumn{1}{c|}{88.19}         & \multicolumn{1}{c|}{\textbf{88.52}} & \textbf{89.07*}                      \\ 
CIFAR-100                 & NS: WRN28-2+WRN40-2, SSS: WRN28-2, LSS: WRN40-2 & \multicolumn{1}{c|}{69.23}         & \multicolumn{1}{c|}{\textbf{69.84}}          & \textbf{71.06*}                      \\ \hline
\end{tabular}
\caption{NoisyStudent comparison results. (NS: NoisyStudent, SSS: SmallerSameSized, LSS: LargerSameSized, \textbf{Bold}: Better than NS, \\ \textbf{*}: Current best result for each dataset)}
\label{Table8a}
\end{table*}

\begin{figure*}[ht!]
\begin{center}
\includegraphics[width=12cm]{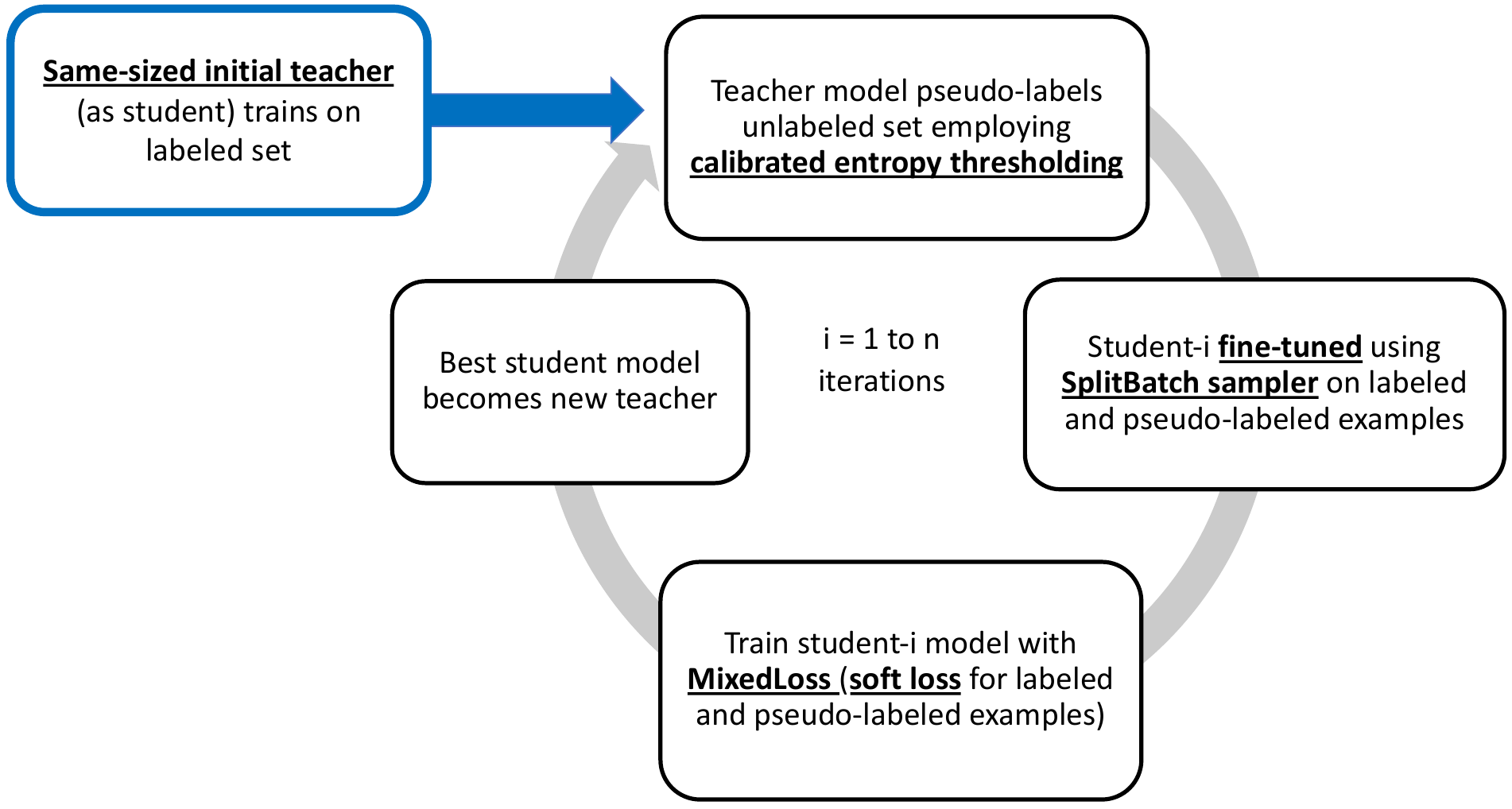}
\end{center}
   \caption{Enhanced self-training (EST) pipeline.}
\label{fig:enhanced}
\end{figure*}

\noindent \textbf{Final Model} \\
We aggregated the best components  from the above experiments, selected sequentially by majority voting on the three datasets, which resulted in selecting (1) Soft Loss, (2) Fine-tuning, (3) Weighted SplitBatch Sampler (w/ associated MixedLoss Function), (5) Calibrated Entropy Thresholding, and (6) Larger Same-Sized Teacher-Student models. The resulting enhanced approach is shown in Fig.~\ref{fig:enhanced}. We refer to this final model as the \textbf{enhanced self-train (EST)} approach.

\subsection{Additional Evaluations}

We next evaluated the generalizability of the EST model with the different labeled/unlabeled subset sizes and larger model architectures. For these experiments, a single run for each experiment is reported.\\

\noindent \textbf{Labeled/Unlabeled Subset Sizes} \\
We evaluated our EST approach on different labeled/unlabeled subset sizes of CINIC-10 with its large number of samples (180K) split into different-sized labeled/unlabeled subsets. First, we examined a small labeled data scenario with 1K labeled examples (100 examples per-class) across increases in unlabeled data (20K-150K). Next, we evaluated a large labeled data scenario with 20K labeled examples (2K examples per-class) across the various unlabeled sets. We applied a 20/80\% mini-batch split in the small label scenario to avoid overfitting when oversampling the smaller labeled set. We used a 50/50\% mini-batch split for the large labeled scenario having many labeled examples to oversample. A balanced loss was used in both scenarios (as in the earlier experiments). We trained all models using a WRN28-8 backbone for the same number of training optimization steps. 

As expected, the results in Table \ref{Table10} show that performance increases as we add more unlabeled examples during training in both scenarios, highlighting the importance of building large unlabeled datasets for semi-supervised learning methods. Our EST approach provided significant improvements in the small labeled scenario while providing slight improvements to the large labeled scenario (as the teacher model already has enough labeled examples for better learning performance).\\


\begin{table}[t]
\centering
\begin{tabular}{|c|cc|}
\hline
\multirow{2}{*}{CINIC-10 Split} & \multicolumn{2}{c|}{Best Top-1 Acc}                   \\
                                & 1K Lab                              & 20K Lab         \\ \hline\hline
Init. Teacher                         & \multicolumn{1}{c|}{53.97}          & 82.55           \\
20K Unlab                       & \multicolumn{1}{c|}{65.82}          & 82.64           \\
50K Unlab                       & \multicolumn{1}{c|}{67.03}          & 82.74           \\
100K Unlab                      & \multicolumn{1}{c|}{67.42}          & 83.24           \\
150K Unlab                      & \multicolumn{1}{c|}{\textbf{67.63*}} & \textbf{83.59*} \\ \hline
\end{tabular}
\caption{Unlabeled dataset size comparison results. (Lab: Labeled dataset size, Unlab: Unlabeled dataset size), (\textbf{Bold}: Best result in table, \textbf{*}: Current best result for each dataset)}
\label{Table10}
\end{table}

\noindent \textbf{Larger Model Architectures} \\
We next evaluated the generalizability of our EST approach using larger models with longer epochs and larger datasets. We evaluated a R34 model on SVHN (previously used R18) with a 20/80\% mini-batch split, a WRN28-8 model on CIFAR-10 and CIFAR-100 (previously used WRN28-2) with a 20/80\% and 40/60\% mini-batch split, respectively, a WRN28-8 model on the CINIC-10 dataset (20K Lab + 150K UnLab) with a 20/80\% mini-batch split (previously used 50/50\% split), and additionally included a R34 model on the TinyImageNet dataset with a 50/50\% mini-batch split. We trained the teacher model for 400 epochs and all student iterations for 200 epochs (previously used 150 epochs) with the SAMix+RandAugmet data augmentation policy for all models/datasets. Interestingly, we found that applying SAMix on SVHN actually helped improve digit classification performance (unlike previous expectations). We compared against the NS approach reimplemented as described in \cite{Xie2020a}  using smaller teacher models as suggested (R18 for SVHN and TinyImageNet, WRN28-2 for CIFAR and CINIC datasets) but the same larger student model sizes as ours. From the results in Table \ref{Tabel11}, we can see that our EST approach outperformed the original NS approach on all the datasets we evaluated.\\

\begin{table}[t]
\centering
\begin{tabular}{|c|cc|}
\hline
\multirow{2}{*}{Datasets} & \multicolumn{2}{c|}{Best Top-1 Acc}          \\
                          & NS                         & EST             \\ \hline\hline
SVHN                      & \multicolumn{1}{c|}{91.65} & \textbf{93.00*} \\
CIFAR-10                  & \multicolumn{1}{c|}{89.15} & \textbf{94.21*} \\
CIFAR-100                 & \multicolumn{1}{c|}{70.53} & \textbf{76.42*} \\
CINIC-10                  & \multicolumn{1}{c|}{83.47} & \textbf{88.59*} \\
TinyImageNet              & \multicolumn{1}{c|}{49.32} & \textbf{52.23*} \\ \hline
\end{tabular}
\caption{EST: Enhanced self-training vs. NS: NoisyStudent best student top-1 accuracy results. (\textbf{Bold}: Best result in table, \textbf{*}: Current best result for each dataset)}
\label{Tabel11}
\end{table}

\subsection{Open Set Data}

Real-world unlabeled data can be from an Open Set that contains data belonging to the target classes (classes from labeled training data) and data from additional non-target classes. The inclusion of non-target class examples in the unlabeled set can degrade SSL performance \cite{Guo2020a}. We propose a basic Open Set recognition technique using contrastive learning to build a feature space for all target classes, where the non-target classes should hopefully be farther away from the target classes. We used SimCLR \cite{Chen2020a} to learn a contrastive feature space from the labeled target data and the unlabeled data (contains target and non-target class examples). We used a validation set to find a mean prototype vector for each known target class and fit a Beta distribution per-class of the distances from labeled target examples to their class prototype. We then filtered out examples expected to be from any non-target class by using a per-class Beta cumulative distribution function (CDF) and a global CDF threshold (learned from validation), where examples having CDF values above the threshold for all classes were considered to be from a non-target class.

We evaluated this method on a custom-built Open Set version of CIFAR-10/100 with labeled and unlabeled subsets. The labeled subset is made up of a Closed Set with 10 target classes (CIFAR-10 subset consisting of 4K images), and the unlabeled subset contains 110 total classes with 10 target classes (CIFAR-10 subset consisting of different 42K images) and 100 non-target classes (CIFAR-100 subset consisting of 42K images). We compared the performance of the NS approach (reimplemented as suggested in \cite{Xie2020a} with a smaller WRN28-2 initial teacher and the same larger WRN28-8 student as EST, thus resulting in a performance decrease) to our EST approach (same-sized WRN28-8 teacher-student). After searching through multiple values, we found that CDF thresholds of 0.85 and 0.9 led to the best Closed Set validation accuracy for NS and EST, respectively. From the results in Table \ref{Tableopen}, we observed that our EST approach performed better than the NS approach showing that our proposed enhancements for handling noisy pseudo-labels extend to Open Set data as well by providing some basic filtering of pseudo-labels belonging to non-target classes. We also found that both the NS and EST approaches had further improvements upon training with our filtered Open Set data, with our EST approach on filtered Open Set data performing the best.

\begin{table}[t]
\centering
\begin{tabular}{|c|c|}
\hline
Experiment Description  & Best Student Top-1 Acc \\
\hline\hline
NS on Labeled Closed Set & 71.82            \\
EST on Labeled Closed Set & 85.83            \\
\hdashline
NS on Open Set & 87.8            \\
EST on Open Set & \textbf{92.62}            \\
\hdashline
NS on Filtered Open Set & 88.14            \\
EST on Filtered Open Set & \textbf{93.12} \\
\hline
\end{tabular}
\caption{Open Set ST results. (EST: Enhanced self-training, NS: Noisy Student, \textbf{Bold}: Best result in table subsection)}
\label{Tableopen}
\end{table}


\section{Conclusion}

We proposed multiple modular enhancements to the standard ST pipeline to alleviate confirmation bias commonly seen in pseudo-labeling SSL methods. We demonstrated an enhanced ST pipeline using confidence calibration, entropy thresholding, and custom sampling techniques to avoid overfitting to noisy pseudo-labeled data. We also demonstrated a basic Open Set recognition technique to augment self-training performance on unlabeled data with novel unseen class distributions. In future work, we plan to directly integrate Open Set recognition capabilities into the ST models and leverage contrastive learning techniques to assist in learning better feature representations to separate known and unknown classes. 

\section{Acknowledgments}
This work was supported in part by the U.S. Air Force Research Laboratory under contract \#GRT00054740. Distribution A: Cleared for Public Release. Distribution Unlimited. PA Approval \#AFRL-2022-5307

{\small
\bibliographystyle{ieee_fullname}
\bibliography{egbib}

\begin{thebibliography}{10}\itemsep=-1pt

\bibitem{Arazo2020a}
Eric Arazo, Diego Ortego, Paul Albert, Noel~E. O’Connor, and Kevin
  McGuinness.
\newblock {Pseudo-Labeling and Confirmation Bias in Deep Semi-Supervised
  Learning}.
\newblock In {\em International Joint Conference on Neural Networks}, 2020.

\bibitem{Athiwaratkun2019a}
Ben Athiwaratkun, Marc Finzi, Pavel Izmailov, and Andrew~Gordon Wilson.
\newblock {There are Many Consistent Explanations of Unlabeled Data: Why you
  should Average}.
\newblock In {\em International Conference on Learning Representations}, 2019.

\bibitem{Chen2020a}
Ting Chen, Simon Kornblith, Mohammad Norouzi, and Geoffrey Hinton.
\newblock {A Simple Framework for Contrastive Learning of Visual
  Representations}.
\newblock In {\em International Conference on Machine Learning}, 2020.

\bibitem{Cubuk2020a}
Ekin~D. Cubuk, Barret Zoph, Jonathon Shlens, and Quoc~V. Le.
\newblock {Randaugment: Practical Automated Data Augmentation with a Reduced
  Search Space}.
\newblock In {\em IEEE/CVF Conference on Computer Vision and Pattern
  Recognition Workshops: Efficient Deep Learning for Computer Vision}, 2020.

\bibitem{Darlow2018a}
Luke Darlow, Elliot Crowley, Antreas Antoniou, and Amos Storkey.
\newblock {{CINIC-10} is not ImageNet or {CIFAR-10}}.
\newblock {\em arXiv preprint arXiv:1810.03505}, 2018.

\bibitem{Deng2009a}
Jia Deng, Wei Dong, Richard Socher, Li-Jia Li, Kai Li, and Li Fei-Fei.
\newblock {ImageNet: A Large-Scale Hierarchical Image Database}.
\newblock In {\em IEEE Conference on Computer Vision and Pattern Recognition},
  2009.

\bibitem{Fawcett2006a}
Tom Fawcett.
\newblock {An Introduction to ROC Analysis}.
\newblock {\em Pattern Recognition Letters}, 2006.

\bibitem{Grandvalet2004a}
Yves Grandvalet and Yoshua Bengio.
\newblock {Semi-Supervised Learning by Entropy Minimization}.
\newblock In {\em Advances in Neural Information Processing Systems}, 2004.

\bibitem{Guo2017a}
Chuan Guo, Geoff Pleiss, Yu Sun, and Kilian~Q. Weinberger.
\newblock {On Calibration of Modern Neural Networks}.
\newblock In {\em International Conference on Machine Learning}, 2017.

\bibitem{Guo2020a}
Lan-Zhe Guo, Zhen-Yu Zhang, Yuan Jiang, Yu-Feng Li, and Zhi-Hua Zhou.
\newblock {Safe Deep Semi-Supervised Learning for Unseen-Class Unlabeled Data}.
\newblock In {\em International Conference on Machine Learning}, 2020.

\bibitem{Haque2021a}
Ayaan Haque, Abdullah-Al-Zubaer Imran, Adam Wang, and Demetri Terzopoulos.
\newblock {Generalized Multi-Task Learning from Substantially Unlabeled
  Multi-Source Medical Image Data}.
\newblock {\em Machine Learning for Biomedical Imaging}, 2021.

\bibitem{He2016a}
Kaiming He, Xiangyu Zhang, Shaoqing Ren, and Jian Sun.
\newblock {Deep Residual Learning for Image Recognition}.
\newblock In {\em IEEE/CVF Conference on Computer Vision and Pattern
  Recognition}, 2016.

\bibitem{Krizhevsky2009a}
Alex Krizhevsky.
\newblock {Learning Multiple Layers of Features from Tiny Images}.
\newblock 2009.

\bibitem{Laine2017a}
Samuli Laine and Timo Aila.
\newblock {Temporal Ensembling for Semi-Supervised Learning}.
\newblock In {\em International Conference on Learning Representations}, 2017.

\bibitem{letiny}
Ya Le and Xuan Yang.
\newblock {Tiny ImageNet Visual Recognition Challenge}.
\newblock 2015.

\bibitem{Li2020a}
Junnan Li, Richard Socher, and Steven~C.H. Hoi.
\newblock {DivideMix: Learning with Noisy Labels as Semi-supervised Learning}.
\newblock In {\em International Conference on Learning Representations}, 2020.

\bibitem{Li2021a}
Siyuan Li, Zicheng Liu, Di Wu, Zihan Liu, and Stan~Z. Li.
\newblock {Boosting Discriminative Visual Representation Learning with
  Scenario-Agnostic Mixup}.
\newblock {\em arXiv preprint arXiv:2111.15454}, 2021.

\bibitem{Liu2022a}
Zhuang Liu, Hanzi Mao, Chao-Yuan Wu, Christoph Feichtenhofer, Trevor Darrell,
  and Saining Xie.
\newblock {A ConvNet for the 2020s}.
\newblock In {\em IEEE/CVF Conference on Computer Vision and Pattern
  Recognition}, 2022.

\bibitem{Mahajan2018a}
Dhruv Mahajan, Ross Girshick, Vignesh Ramanathan, Kaiming He, Manohar Paluri,
  Yixuan Li, Ashwin Bharambe, and Laurens van~der Maaten.
\newblock {Exploring the Limits of Weakly Supervised Pretraining}.
\newblock In {\em European Conference on Computer Vision}, 2018.

\bibitem{Motlagh2021a}
Nicholas~Kashani Motlagh, Aswathnarayan Radhakrishnan, Jim Davis, and Roman
  Ilin.
\newblock {A Framework for Semi-Automatic Collection of Temporal Satellite
  Imagery for Analysis of Dynamic Regions}.
\newblock In {\em IEEE/CVF International Conference on Computer Vision
  Workshops: Learning to Understand Aerial Images}, 2021.

\bibitem{Naeini2015a}
Mahdi~Pakdaman Naeini, Gregory~F. Cooper, and Milos Hauskrecht.
\newblock {Obtaining Well Calibrated Probabilities using Bayesian Binning}.
\newblock In {\em AAAI Conference on Artificial Intelligence}, 2015.

\bibitem{Netzer2011}
Yuval Netzer, Tao Wang, Adam Coates, Alessandro Bissacco, Bo Wu, and Andrew Ng.
\newblock {Reading Digits in Natural Images with Unsupervised Feature
  Learning}.
\newblock In {\em Advances in Neural Information Processing Systems}, 01 2011.

\bibitem{Chapelle2009a}
Chapelle Olivier, Scholkopf Bernhard, and Zien Alexander.
\newblock {Semi-Supervised Learning}.
\newblock {\em IEEE Transactions on Neural Networks}, 2009.

\bibitem{Ouali2020a}
Yassine Ouali, Céline Hudelot, and Myriam Tami.
\newblock {An Overview of Deep Semi-Supervised Learning}.
\newblock {\em arXiv preprint arXiv:2006.05278}, 2020.

\bibitem{Radhakrishnan2019a}
Aswathnarayan Radhakrishnan, Jamie Cunningham, Jim Davis, and Roman Ilin.
\newblock {A Framework for Collecting and Classifying Objects in Satellite
  Imagery}.
\newblock In {\em Advances in Visual Computing}. Springer International
  Publishing, 2019.

\bibitem{Riloff1996a}
Ellen Riloff.
\newblock {Automatically Generating Extraction Patterns from Untagged Text}.
\newblock In {\em Proceedings of the Thirteenth National Conference on
  Artificial Intelligence - Volume 2}. AAAI Press, 1996.

\bibitem{Riloff2003a}
Ellen Riloff and Janyce Wiebe.
\newblock {Learning Extraction Patterns for Subjective Expressions}.
\newblock In {\em Proceedings of the 2003 Conference on Empirical Methods in
  Natural Language Processing}. Association for Computational Linguistics,
  2003.

\bibitem{Sajjadi2016a}
Mehdi Sajjadi, Mehran Javanmardi, and Tolga Tasdizen.
\newblock {Regularization with Stochastic Transformations and Perturbations for
  Deep Semi-Supervised Learning}.
\newblock In {\em Advances in Neural Information Processing Systems}, 2016.

\bibitem{Scudder1965a}
H. Scudder.
\newblock {Probability of error of some adaptive pattern-recognition machines}.
\newblock {\em IEEE Transactions on Information Theory}, 1965.

\bibitem{Srivastava2014a}
Nitish Srivastava, Geoffrey Hinton, Alex Krizhevsky, Ilya Sutskever, and Ruslan
  Salakhutdinov.
\newblock {Dropout: A Simple Way to Prevent Neural Networks from Overfitting}.
\newblock {\em Journal of Machine Learning Research}, 2014.

\bibitem{Su2021a}
Jong-Chyi Su, Zezhou Cheng, and Subhransu Maji.
\newblock {A Realistic Evaluation of Semi-Supervised Learning for Fine-Grained
  Classification}.
\newblock In {\em IEEE/CVF Conference on Computer Vision and Pattern
  Recognition}, 2021.

\bibitem{Tajbakhsh20201a}
Nima Tajbakhsh, Laura Jeyaseelan, Qian Li, Jeffrey~N. Chiang, Zhihao Wu, and
  Xiaowei Ding.
\newblock {Embracing Imperfect Datasets: A Review of Deep Learning Solutions
  for Medical Image Segmentation}.
\newblock {\em Medical Image Analysis}, 2020.

\bibitem{Tarvainen2017a}
Antti Tarvainen and Harri Valpola.
\newblock {Mean Teachers Are Better Role Models: Weight-Averaged Consistency
  Targets Improve Semi-Supervised Deep Learning Results}.
\newblock In {\em Advances in Neural Information Processing Systems}, 2017.

\bibitem{Xie2020a}
Qizhe Xie, Minh-Thang Luong, Eduard Hovy, and Quoc~V. Le.
\newblock {Self-Training with Noisy Student improves ImageNet Classification}.
\newblock In {\em IEEE/CVF Conference on Computer Vision and Pattern
  Recognition}, 2020.

\bibitem{Yang2022a}
Fan Yang, Kai Wu, Shuyi Zhang, Guannan Jiang, Yong Liu, Feng Zheng, Wei Zhang,
  Chengjie Wang, and Long Zeng.
\newblock {Class-Aware Contrastive Semi-Supervised Learning}.
\newblock In {\em IEEE/CVF Conference on Computer Vision and Pattern
  Recognition}, 2022.

\bibitem{Yarowsky1995a}
David Yarowsky.
\newblock {Unsupervised Word Sense Disambiguation Rivaling Supervised Methods}.
\newblock In {\em Proceedings of the 33rd Annual Meeting on Association for
  Computational Linguistics}. Association for Computational Linguistics, 1995.

\bibitem{Zagoruyko2016WideRN}
Sergey Zagoruyko and Nikos Komodakis.
\newblock {Wide Residual Networks}.
\newblock {\em arXiv preprint arXiv:1605.07146}, 2016.

\bibitem{Zhai2022a}
Xiaohua Zhai, Alexander Kolesnikov, Neil Houlsby, and Lucas Beyer.
\newblock {Scaling Vision Transformers}.
\newblock In {\em IEEE/CVF Conference on Computer Vision and Pattern
  Recognition}, 2022.

\bibitem{Zhang2018a}
Hongyi Zhang, Moustapha Cisse, Yann~N. Dauphin, and David Lopez-Paz.
\newblock {mixup: Beyond Empirical Risk Minimization}.
\newblock In {\em International Conference on Learning Representations}, 2018.

\bibitem{Zhu2005a}
Xiaojin Zhu.
\newblock {Semi-Supervised Learning Literature Survey}.
\newblock Technical report, Computer Sciences, University of Wisconsin-Madison,
  2005.

\end{thebibliography}
}

\end{document}